\documentclass[pmlr]{jmlr}

\begin{document}
% Heading arguments are {volume}{year}{pages}{submitted}{published}{author-full-names}

% \jmlrheading{1}{2000}{1-48}{4/00}{10/00}{Mayank Kejriwal and Shilpa Thomas}

% Short headings should be running head and authors last names

% \ShortHeadings{GNOME}{Kejriwal and Thomas}
% \firstpageno{1}

\title{Generating Novelty in Open-World Multi-Agent Strategic Board Games}

\author{\Name{Mayank Kejriwal} \Email{kejriwal@isi.edu} \\
       \addr Information Sciences Institute\\
       University of Southern California\\
       Marina del Rey, CA 90292, USA
       \AND
       \Name{Shilpa Thomas} \Email{shthomas@isi.edu} \\
       \addr Information Sciences Institute\\
       University of Southern California\\
       Marina del Rey, CA 90292, USA}

% \editor{Leslie Pack Kaelbling}

\maketitle

\begin{abstract}%   <- trailing '%' for backward compatibility of .sty file
 We describe GNOME (Generating Novelty in Open-world Multi-agent Environments), an experimental platform that is designed to test the effectiveness of multi-agent AI systems when faced with \emph{novelty}. GNOME separates the development of AI gameplaying agents with the simulator, allowing \emph{unanticipated} novelty (in essence, novelty that is not subject to model-selection bias). Using a Web GUI, GNOME was recently demonstrated at NeurIPS 2020 using the game of Monopoly to foster an open discussion on AI robustness and the nature of novelty in real-world environments. In this article, we further detail the key elements of the demonstration, and also provide an overview of the experimental design that is being currently used in the DARPA Science of Artificial Intelligence and Learning for Open-World Novelty (SAIL-ON) program to evaluate external teams developing novelty-adaptive gameplaying agents. 
\end{abstract}


\section{Introduction}

% FAQs.
% Q: If my paper is accepted, will it be published by the JMLR?
% A: No, accepted papers will be published as part of the Proceedings track associated with JMLR,see
% http://proceedings.mlr.press/v123/
% for details
% Q: Is it possible to include appendices in the manuscript?
% A: Yes, as long you keep it to a reasonable size
% Q: Are papers expected to have the structure of a "standard" scientific paper.
% A: Yes, please note the papers will undergo a review process that will evaluate the scientific /technological contribution of the paper itself. Please make sure to include a review of relevantrelated work, and highlight the contributions of the paper, together with their potential impact.

Multi-agent gameplaying is a difficult research challenge in AI [\cite{diplomacy}], even with recent advances in deep reinforcement learning [\cite{alphago}, \cite{alphazero}]. In multi-agent (and typically, stochastic) board games of strategy (such as Monopoly, Risk, and also Diplomacy), the decision space can be vast, and the game environment contains both relevant and irrelevant elements. While recent work on agents has illustrated enormous promise, the prototypical agent is developed with the understanding of a `default' game\footnote{Even if the rules are unknown.} that does not change during play. Therefore, the only variance is due to stochasticity (such as die rolls) or due to decisions made by other agents. 

While this is a useful abstraction, real-world environments can, and do, change. Furthermore, it is not always possible in advance to anticipate such changes. Arguably, an agent that exhibits a true understanding of the general principles in a given domain (e.g., chess-playing) should be able to detect and adapt to novelties in the domain. This is a new area of research for which neither evaluation platforms (or simulators) nor a mapped-out research agenda exists. While experimenters can always inject ad-hoc novelties into a domain environment, what is necessary for systematic and `double-blind' experiments (where novelty-adaptive agents are developed by teams independent of the team that is actually injecting the novelties and doing the experimenting) is a robust simulator that permits injection of novelty at multiple levels (including new objects, attributes and representation). To support a fully developed novelty-centric research agenda, the simulator should also be capable of computing metrics that indicate whether an agent is successfully reacting to novelty.

We address these challenges by describing GNOME (Generating Novelty in Open-world Multi-agent Environments), a simulator for injecting (and evaluating AI agents on) open-world novelty within the context of a multi-agent game such as Monopoly or Poker. GNOME is funded under the DARPA SAIL-ON program\footnote{\url{https://www.darpa.mil/program/science-of-artificial-intelligence-and-learning-for-open-world-novelty}}. SAIL-ON is tasked with researching the underlying scientific principles and AI algorithms necessary for training agents that act effectively in novel situations that occur in \emph{open worlds}. AI agents must start reacting as soon as the novelty presents itself, and are not allowed to go `offline' to re-train or to observe many instances. GNOME provides an advanced simulator that evaluates candidate agents (usually developed by other organizations and teams) through generation and combination of novelties of escalating difficulty. Although GNOME is eventually expected to support several multi-agent board games of strategy, we use the classic game of Monopoly to illustrate its facilities in this article. We also describe the elements of the system that was demonstrated recently at the NeurIPS conference in late 2020 wherein participants were able to go to a Web GUI, inject a wide variety of novelties, and experience for themselves on a 2D gameboard how pre-programmed agents reacted in the face of that novelty.

\section{Background and Related Work}\label{rw}

While novelty is an important subject of study in both philosophy [\cite{phil1}, \cite{phil2}], and cognitive science [\cite{simon1971}, \cite{chu2011}], it has only received indirect attention in AI. \cite{simon1971} theorized that problem-solving was a primary mechanism through which humans responded to novelty. Of course, both early and modern work on machine learning extensively addressed the notion of generalization [see, for example, \cite{ML1}, \cite{ML2}, \cite{ML3}, \cite{ML4}, \cite{ML5}], but the `novelty' embodied in the `test' data was (at least originally) understood to have been sampled from the same distribution as the `training' distribution that was  used to infer the parameters of the model. 

Two lines of work that seem to have treated novelty as first-class citizens are concept drift and anomaly detection [\cite{conceptdrift}, \cite{anomalydetection}], with applications in time series analysis and fraud detection [\cite{fraud}, \cite{TS}]. A survey of novelty detection was provided by \cite{pimentel2014}. However, far less research has been conducted on novelty that is more \emph{structural} than \emph{distributional}. In both symbolic and agent-centric domains, it is this kind of novelty that is central. Observations are few and discrete, and cannot easily be interpreted using a purely quantitative or mathematical framework. Eventually, as the SAIL-ON program progresses over the next several years, more structural theories of novelty detection may arise. Our goal in this paper is not to present such a theory, or even new algorithmic techniques to react to (or detect) novelty. Rather, the goal behind GNOME is to enable research on these problems by providing a highly customizable simulator, whereby agent development and novelty scripting (and injection) happen relatively independently. 

Owing to a resurgence in the popularity of reinforcement learning for training powerful gameplaying agents, community-driven frameworks have been developed and released, an excellent example being the OpenAI Gym, which is described as both a `toolkit for reinforcement learning' and as a `growing collection of
benchmark problems that expose a common interface' [\cite{openaigym}]. OpenAI Gym has been very influential in enabling reinforcement learning research [see, for example, \cite{ho2016generative}, \cite{pathak2017curiosity}, \cite{arulkumaran2017deep}], and GNOME's APIs and design structure were certainly inspired by it. GNOME's novelty generator is unique, however, and with no obvious analog in the OpenAI Gym environment. We also note that GNOME is not designed to test new reinforcement learning algorithmic innovations or systems \emph{per se}, since it is not completely settled that RL is necessarily the best paradigm for reacting to unanticipated novelty in near real-time. In fact, two of three teams that have been evaluated using GNOME in the SAIL-ON program did not use RL but instead, drew on techniques like probabilistic reasoning and even planning. This may change in the future, of course, with new advancements, but GNOME is agnostic to the algorithms that power the agent's logic at the backend. The only requirement is  to adhere to the requirements of the interface between the agent logic and the simulator.   

Recent strides in gameplaying and AI have been exciting, especially given well-publicized successes such as work by \cite{alphago}, and \cite{alphazero}. Although AI programs have been developed for video games [such as the classic Atari games and Starcraft; see work by \cite{atari} and \cite{starcraft1}], Monopoly and other such games (i.e. multi-agent \emph{board} games) have been relatively unexplored. However, building good AI agents to play games like Starcraft, offers the closest approximation to researching virtual AI agents that can navigate open worlds, and is one sign of progress toward more general artificial intelligence. Rare exceptions include Poker and No-press Diplomacy [\cite{diplomacy}, \cite{poker1}], though little of the software is openly available for other researchers to experiment with, and the environments do not seem to account for injection of novelties. The proposed GNOME platform can be used to advance research in developing, not only agents that can play strategic board games with a mix of agents and board elements, but also agents that can react to novelty in such domains. 

\section{Preliminaries: The Game of Monopoly}
\begin{figure}
\centering
\includegraphics[width=15cm]{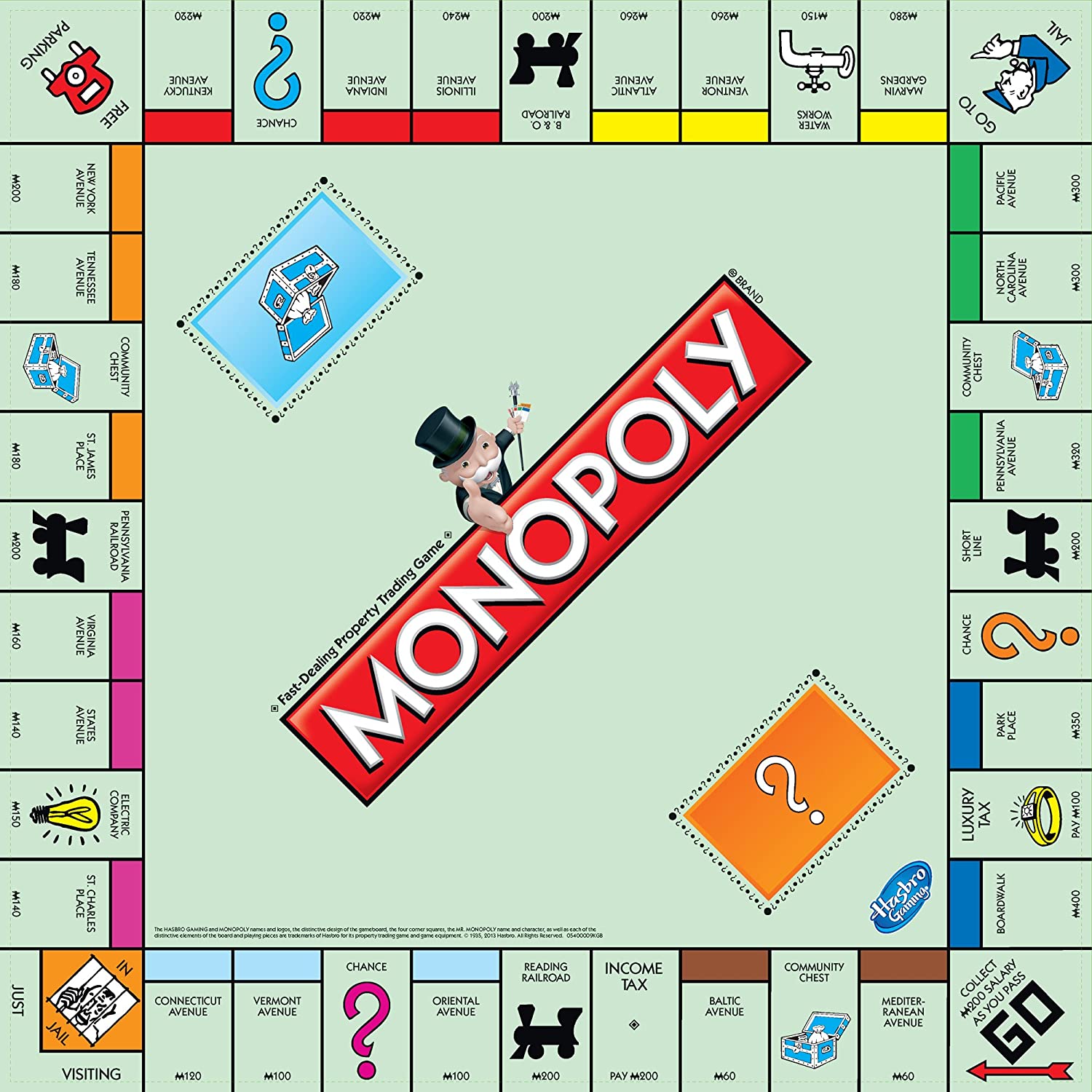}
\caption{An illustration of a typical US-version Monopoly gameboard. Original source: \url{https://www.amazon.com/Hasbro-Monopoly-Replacement-Board/dp/B017MNUCXC}.}
\label{fig:board}
\end{figure}
Monopoly is a multi-agent boardgame involving four players who take turns and make decisions after rolling two unbiased dice. A typical Monopoly board focuses on the real estate market, with the name of the game referring to the business practice of `cornering the market' by becoming the single dominant commercial provider. The physical board is square in shape, with the majority of \emph{slots}\footnote{A slot is a `position' on the gameboard that (one or more players) can \emph{occupy} at any given point of time.} representing `colored' real-estate properties originally owned by a bank, but that are purchaseable, sellable and tradable between the players. Figure \ref{fig:board} illustrates the layout. As mentioned above, each purchasable property (not including \emph{railroads} and \emph{utilities}) is associated with a color e.g., Mediterranean Avenue is associated with the brown color. Players establish monopolies by owning all properties within a color group. Once a monopoly is established, a player can `improve' a property by setting up increasing numbers of houses and hotels on the property. The rents increase rapidly as the property is monopolized. 

The game is approximately zero-sum:  there is only one winner, and a winner emerges after other players gradually become bankrupt, which typically happens when they are subject to the high rents of other players' monopolies. Monopoly-like situations can also arise when players buy more railroads (or to a lesser extent, utilities like Electric Company) as the payment due on landing on a railroad is proportional to the number of railroads owned by the payment-receiving player. Players do get a `Go' increment (in the amount of \$200 in the default game) each time they complete a round trip, but rents due on monopolized and improved properties are so high that this increment is not enough to sustain for long a player who does not also have a revenue-generating source of their own.

% As the name of the game suggests, the objective of the game is to bankrupt all other players and establish (by virtue of bankrupting all other players) a `complete' monopoly. Usually, this can only be accomplished by steadily acquiring entire color groups and forcing other players into bankruptcy due to the rent charges. Since it is highly unlikely that a single player will be able to acquire an entire color group by chance alone, trading and cooperation are important, as are both short-term and long-term decision horizons.

The software and research described in this paper is based on the US version of Monopoly, which has a total of 40 slots\footnote{\url{https://en.wikipedia.org/wiki/Template:Monopoly$\_$board$\_$layout}}. Prices and rents of all properties may be found in publicly available descriptions of the game\footnote{\url{http://www.falstad.com/monopoly.html}}. 
Along with the two dice, the game involves a further element of stochasticity; namely, through 16 \emph{chance}\footnote{\url{https://monopoly.fandom.com/wiki/Chance}} and 16 \emph{community chest}\footnote{\url{https://monopoly.fandom.com/wiki/Community$\_$Chest}} cards. When a player lands on a Chance/Community Card slot\footnote{Indicated on the board with a question mark on the slot.} that requires them to pick a card from one of these packs, they must do so and follow the instructions stated on the card, the effects of which may benefit, be costly to, or (rarely) be neutral to that player. After following the instructions, the player has to return the card to the appropriate deck, the only exception being the `Get out of Jail Free' card, which can be retained and used by the player in the future to get out of jail for free (in the event that the player actually ends up in jail).

% \begin{figure*}
% \centering
% \includegraphics[width=\textwidth,height=5cm]{figure/cc_chance_cards.png}
% \caption{Examples of chance and community chest cards. Original Source: \cite{cards}.}
% \label{fig:card}
% \end{figure*}

\section{Novelty: Working Definition, Categories and Examples}\label{novelty}

Even as a computational concept, novelty can prove to be surprisingly difficult to operationalize (or even define). The original broad agency announcement (BAA) of the SAIL-ON program sought to define novel situations as `those that violate implicit or explicit assumptions about an agent's model of the external world, including other agents, the environment,
and their interactions.' While this is a good definition, it encompasses a very broad space of novelties. One of the ways in which we have tried to constrain the space is by agreeing on a set of \emph{novelty categories}. For example, an \emph{attribute} novelty is one where (as the name suggests) the attribute of a class of objects can change e.g., the blue-colored `Boardwalk' property on a US Monopoly board might now be colored lime-green (which  did not exist as a color on the board before). The change may or may not have substantial impact. For the example above, the change is more substantial than it seems, for one could now purchase Boardwalk and start `improving' it (by building houses and hotels) without needing to acquire any other properties. The reason is that, in Monopoly, all properties of a given color must be owned by a single player before any one of them can be improved. Since Boardwalk now falls in a color-class of its own, it can be improved without further acquisition. This is also true for the lone blue-colored property (Park Place) now left on the board. In contrast, prior to the novelty being injected, both Board Walk and Park Place (which are the most expensive properties on the board) would have to be acquired by a player before either of them could be improved. Any agent that has sufficient understanding of Monopoly would naturally try to take advantage of this novelty. On the other hand, if both Boardwalk and Park Place had been turned to lime-green, the novelty would hardly have qualified as substantial. A Monopoly-playing agent with no novelty adaptation capabilities could continue to play this `new' version of the game without trouble (assuming it was syntactically robust to the fact that a new color has been introduced to the board and an old color has been removed). 

In Monopoly, attribute novelties are particularly common as many locations have a range of customizable attributes, including colors, rents and prices. Another category of novelty is a \emph{class novelty}, defined intuitively as the introduction of `previously unseen classes of objects or entities.' An example of an unseen class of objects can be a new kind of accommodation or improvement (in addition to the houses and hotels available in the default game). An example of a new entity from an existing class is another dice, a novelty that we consider in the demonstration version of the simulator, as we subsequently discuss. We could even introduce a `biased' dice, which is a class novelty, though it is less clear if this is an entity from an `unseen' class of objects. For practical purposes, the distinction is not an important one. 

A third interesting category of novelty is a \emph{representation novelty}, defined as a `change in how entities and features are specified, corresponding to a transformation of dimensions or coordinate system, not necessarily spatial or temporal.' These tend to be most relevant for visual, rather than symbolic, domains. For example, if the game being played was Angry Birds, then turning the frame upside down would be a representation novelty. Since our system is designed for symbolic, rather than visual, gameplaying, representation novelty is less relevant. An obvious example would be to change the layout of the board by scrambling the locations. 

As described in Section \ref{rw}, the closest that the AI community has come to building virtual agents that can operate in open worlds is witnessed in recent advances in playing `open-world' games such as Starcraft [\cite{agent2}]. However, it is not clear how one could build similar agents that are able to react equally well when games such as Chess or Go, with prescribed rules, are subject to the kinds of unanticipated novelty that frequently occur even in everyday situations (e.g., if the plumbing stops working, or the car breaks down). The guiding hypothesis about the open world is that not every situation is knowable in advance; hence, without a deeper understanding of \emph{general} principles of the domain, and without resorting to re-training, an agent trained for the `default' version of the domain would likely not be able to deal with more advanced novelties. 
% In the next section, we describe how GNOME enables research in this area in the  multi-agent gameplaying domain of Monopoly. While beyond the scope of this article, we also comment briefly on whether the hypothesis stated above about agent adaptation to challenging novelties has (preliminary) empirical support.  
 
\section{Generating Novelty in Open-world Multi-agent Environments (GNOME): System Demonstration at NeurIPS 2020}
% {\bf Demonstration Plan and Audience Interaction.} 
We illustrate the key elements of GNOME's live demonstration in NeurIPS in this section. We set up, and ran, the simulator on a cloud server that was accessible though a Web browser. To ensure a smooth initial experience with light cognitive effort, we designed a roughly linear workflow: 
% \begin{figure}
% \centering
% \includegraphics[width=5.6in, height=7.0in]{fig1}
% \caption{The Web GUI and elements of audience interaction.}\label{fig1}
% \end{figure}
\begin{figure}
\centering
\includegraphics[width=5.6in]{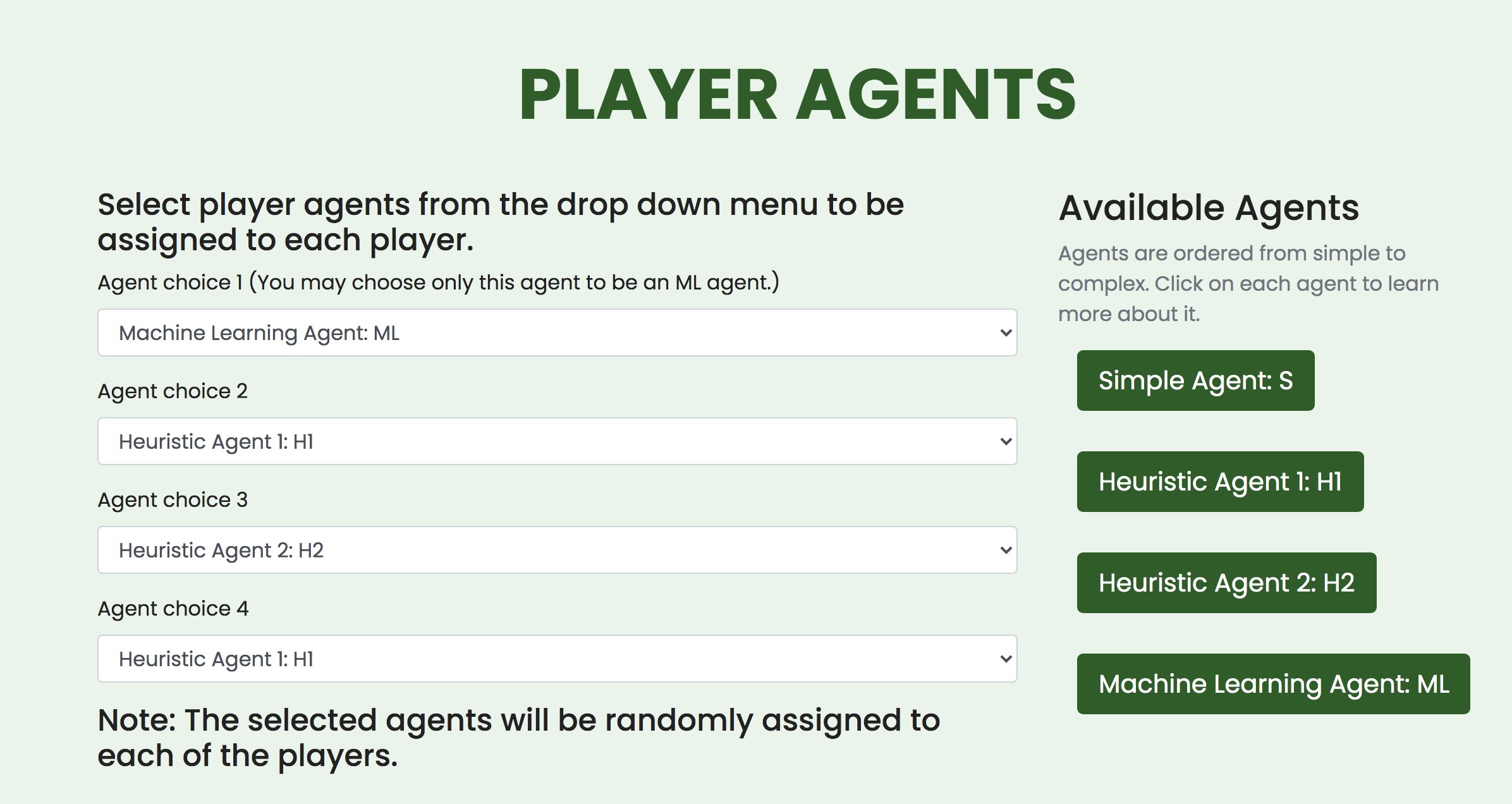}
\caption{The first step (agent combination selection) in a linear demonstration workflow for GNOME.}\label{step1}
\end{figure}

\begin{figure}
\centering
\includegraphics[width=5.6in]{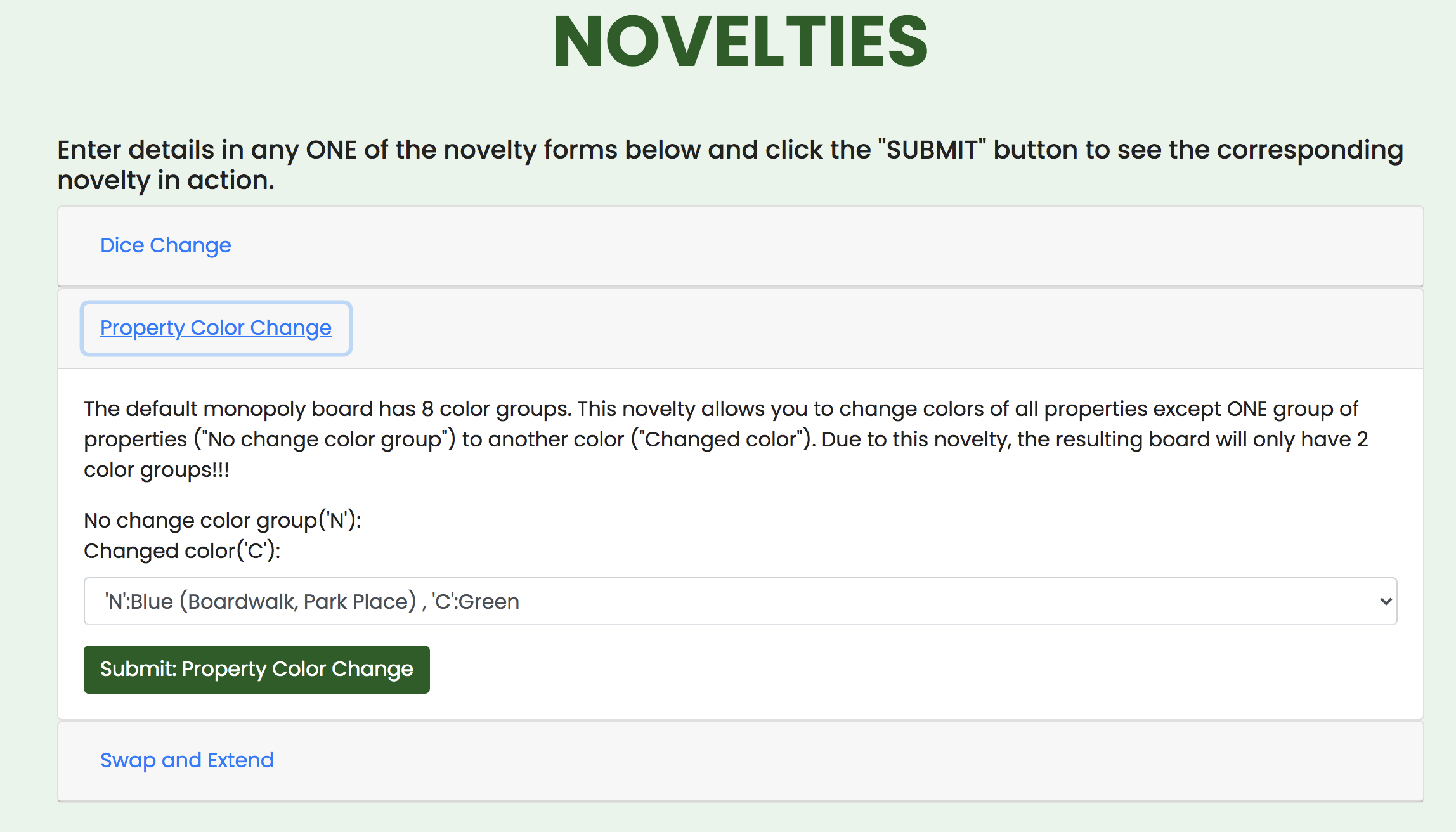}
\caption{The second step (novelty injection) in a linear demonstration workflow for GNOME.}\label{step2}
\end{figure}

\begin{figure}
\centering
\includegraphics[width=5.6in]{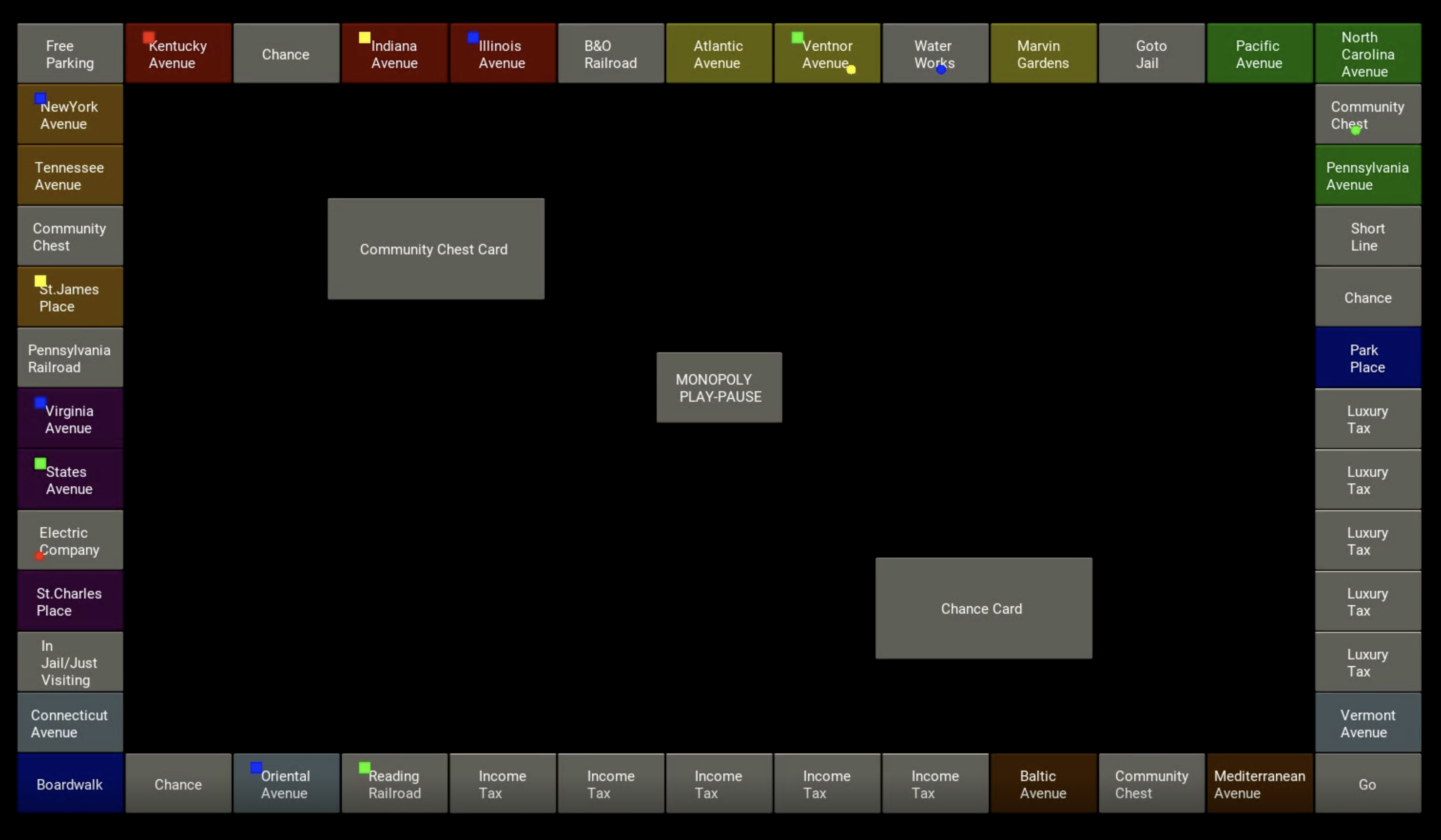}
\caption{An example of Monopoly board with a `Swap and Extend' novelty instance injected into it. Both the luxury and income tax locations now extend to 5 consecutive slots each.}\label{step2-1}
\end{figure}

\begin{enumerate}
    \item \textbf{Step 1:} As a first step (Figure \ref{step1}), the attendee has to select the combination of the four agents that will play the Monopoly game. We provided a library of pre-programmed agents, including a `simple' agent that makes minimal decisions and desires neither to trade nor improve properties by setting up houses or hotels. Furthermore, if the agent runs out of cash (maybe because of a high rent payment) in a given round, it would declare bankruptcy rather than attempt to sell off, or mortgage, its properties to stay in the game. Since this is an extremely simple agent, we also provide  two `heuristic' agents (H1\footnote{H1 is capable of making decisions on how and when to set up houses and hotels on monopolized color groups, and is also capable of trading. It is also able to better address the situation when it is perilously close to running out of cash. For example, whenever the player associated with this agent is low on cash, the agent can make a trade offer by offering one of its properties to another player in return for cash. It may also mortgage or sell properties and improvements.} and H2\footnote{H2 builds on H1 primarily through more sophisticated trading and monopoly-acquisition strategies. Specifically, the agent is capable of making 2-way trade offers where it can both offer and request properties simultaneously, rather than deal with cash. It can also roll out trade offers simultaneously to multiple players.}) that are progressively more sophisticated, although the strategies are in the form of hard-coded rules. Finally, the ML agent (arguably the most sophisticated, but also the least predictable) is trained using reinforcement learning. Since the action and state space in Monopoly is very large, we only trained for some decisions (such as whether to buy a property on which the agent has landed); for other decisions, the agent relies on heuristics. Hence, it is more appropriate to state that this is not a pure ML agent, but is hybrid. In Figure \ref{step1}, the user has selected two H1 agents, an H2 agent and an ML agent. Note that it is possible to have two or more players have the same `agent', since the agent only provides the logic and strategies for playing the game. Each player still maintains its own game state. This facility can sometimes lead to interesting behavior\footnote{For example, we discovered in some of our experiments that the performance of an agent can depend on whether another instance of that agent is also playing the game. Counter-intuitively, the \emph{average} performance of H2 can decline when more than one instance is playing (and because this is zero-sum game, the average performance of another `weaker' player, such as instantiated with a simple agent, would become better) rather than a single instance. Once novelty is injected, the situation becomes even more interesting. This makes GNOME a good experimental platform for exploring such game-theoretic hypotheses, especially in the face of novelty. Several such experimental runs are ongoing, with agents provided by external organizations.}.   
    \item \textbf{Step 2:} Once the agents have been selected, the user can select the novelty to inject into the game (Figure \ref{step2}). While GNOME can currently support hundreds of different novelties, which could be injected in combinations, we provided a smaller set of interesting and intuitive options in the demonstration, both to avoid choice paralysis and for efficiency reasons. Furthermore, in the demo, the user was not allowed to inject more than one novelty at a time. A \emph{Dice Novelty} is the easiest to understand and allows the user to increase the number of dice from 2 to 3-5. Although this may seem like a minor change, it turns out that the novelty reduces the expected revenue from a monopoly, since the probability that a user will land consecutively on a monopolized property in the same color group (e.g., Oriental Avenue and Connecticut Avenue in Figure \ref{fig:board}) becomes lower. In some cases, it goes from an already low probability\footnote{In the `default' game, with probability 1/36 (both dice have to turn up 1), a player who is on Park Place may land on Boardwalk in the next move. In the new game, with 3 or more dice, the probability is 0.} to 0. In contrast, the expected revenue from owning all four railroads is significant. Therefore, a \emph{novelty-adaptive} agent would be able to take advantage of even this situation to best its opponents. 
    
    The second category of novelties is called \emph{Property Color Change}. Note that the default Monopoly board has 8 color groups (Figure \ref{fig:board}). As the description states in Figure \ref{step2}, this novelty allows the user to change the colors of all properties to a single color, except \emph{one} group of properties, which stay at their original color. Due to this novelty, the resulting board will only have two color groups for real-estate properties (and hence, only two possible monopolies, as opposed to 8 in the default game). Other properties and slots remain unchanged. With this novelty, an agent has limited opportunities to win the game. It can either try to acquire as many railroads as possible (the rents of which climb steeply, the more railroads that the player owns) or, with some luck and ingenuity, monopolize the properties in the \emph{unchanged} color group. This way, it becomes the dominant revenue generator, and in a few round trips, can bankrupt everyone else who lands on the monopolized properties.
    % \footnote{It is worth noting here that, upon completing a round-trip and passing Go, each player gets 200\$ from the bank. Without acquiring monopolies (or enough railroads), this is the largest source of revenue for a player, but it is not enough to save them from bankruptcy if they land even two or three times (in a reasonable range of round-trips) on any property that has been improved enough.}
    
    The third category of novelties, called \emph{Swap and Extend}, injects a representation novelty into the gameboard by increasing the number of slots for a selected property. The board that would be generated if the two tax locations were extended to 5 slots, for example, is shown in Figure \ref{step2-1}. Any player that wants to survive long enough in the game, even with adequate revenue sources, would want to preserve cash, since the tax is 200\$. With the slot extensions, it can be incurred multiple times in a round-trip leading to many hundreds of dollars of losses. 
    
    Interestingly, the outcome and strategies can be asymmetric. While tax collection applies to all players equally, a different play is required if a real-estate location is extended in the same way. The expected revenue from that property increases by a multiple; hence, it is in a smart player's best interest to acquire and monopolize that property, even if it requires trading at a steep premium. Once again, we emphasize that agents developed in the overall SAIL-ON program that are trained to play in the GNOME simulator have \emph{not} been trained on such novelties, and have to adapt to them in real time (i.e., in the span of just a few games). While we release such novelties to the teams as examples and as enablers for research, development and prototyping, the novelties used in the evaluations are designed to be unanticipated and open-world. Regardless of how well they play in the `default' game, only agents that have grasped the principles of the domain in a sufficiently general way are expected to adapt robustly to such novelties. 
   
    \item \textbf{Step 3:} Once the user has decided upon which novelty to inject, the actual gameplay can be visualized in a rudimentary GUI that we developed in a new tab (Figure \ref{step3}). Note that the GUI is not designed for interactive gameplaying, and agents developed to work with GNOME typically use logs produced by the simulator, rather than the GUI, to understand the effects of the simulator. Nevertheless, the GUI provides a lot of information that can be visualized as a short video. In the figure, we used an instance of the \emph{Property Color Change} novelty by changing all property colors (except Boardwalk and Park Place) to green. Because none of the current agents used in the demonstration, including the ML agent, is novelty-adaptive in real time, the game could enter a state of non-termination depending on whether a single player was lucky enough to acquire both Park Place and Boardwalk, improve those properties and bankrupt other players that were \emph{unlucky} enough to land on them a few times. When two different players acquire each of Boardwalk and Park Place, it is usually the case that they both attempt trades with each other when possible, but neither is willing to give up their property due to its potential for a monopoly. This is what we observe for most game runs. Since each player gets a `Go' increment of 200\$ each round, the game tends to go on forever for specific novelties such as this. In contrast, dice novelties and swap-and-extend novelties run almost as fast, or even faster, than a default game run, and terminate within 200-500 board round-trips per player.  In the GUI, we show all essential information required to assess gameplay (including details such as the amount of cash each player is holding, on the side). The `colored circles' on the board represent the players and their current positions, while the colored squares are used to indicate ownership by the corresponding player. The frames can be buffered at a very fast speed if necessary, and even be saved as a video.   
\end{enumerate}
\begin{figure}
\centering
\includegraphics[width=5.6in]{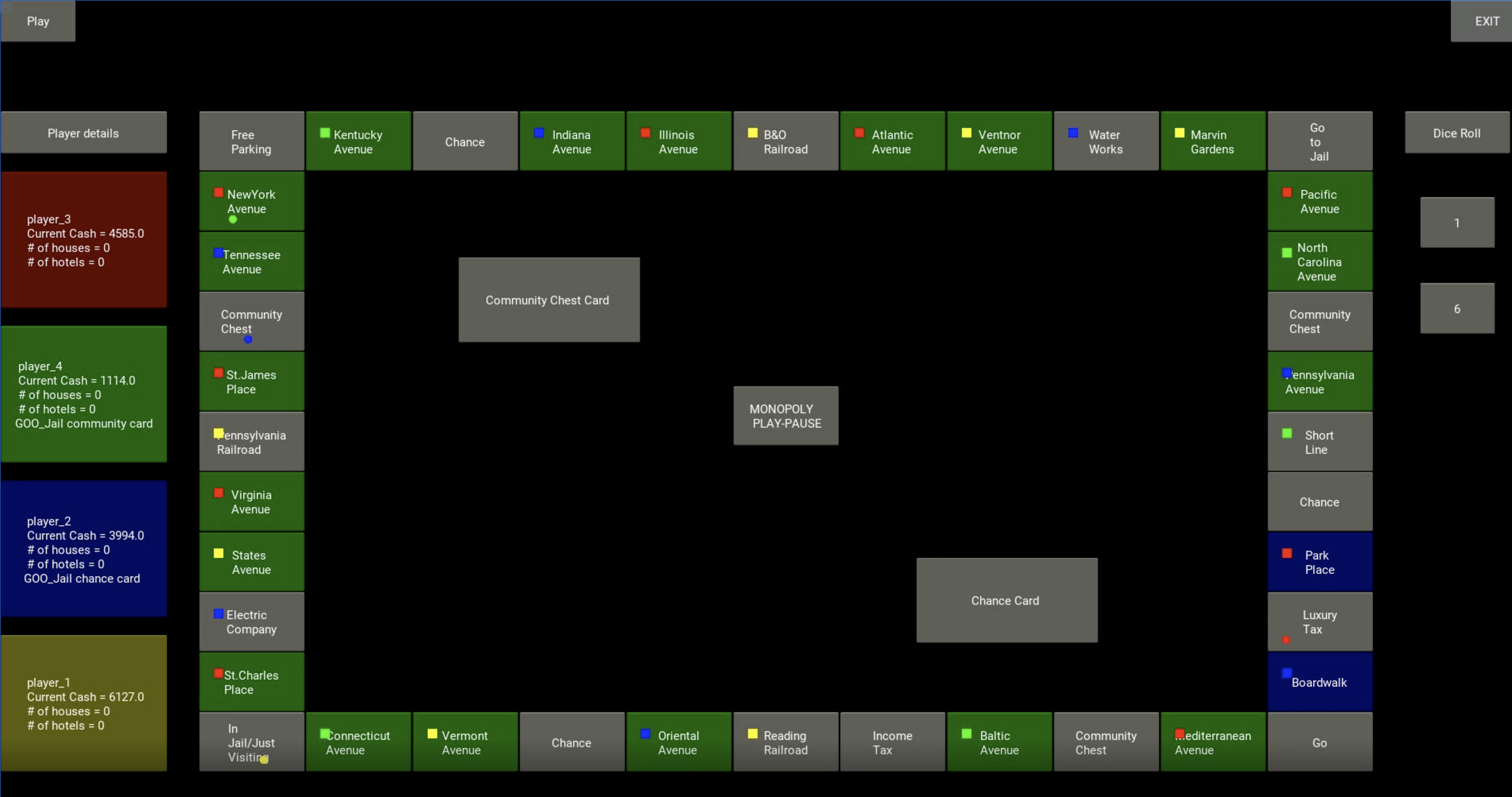}
\caption{The third step (actual gameplay after novelty injection) in a linear demonstration workflow for GNOME. For this illustration, we used an instance of the `Property Color Change' novelty, where colors of Boardwalk and Park Place are preserved, but every other property turns `green'.}\label{step3}
\end{figure}

While the demonstration and facilities described above serve as a good linear workflow, the participant can then choose to explore further by modifying agent combinations or the injected novelty (or both) and re-generating the gameplay. Because gameplay opens in a different tab and we render it as a video by generating it from the logs, it is possible to have several different gameplays open in multiple tabs, and compare and contrast how the game evolves when subject to different experimental conditions. 

\section{Using GNOME for Novelty-driven Experiments}
The GNOME simulator is publicly available\footnote{The homepage is \url{https://usc-isi-i2.github.io/gnome/}, and the software is hosted at \url{https://github.com/mayankkejriwal/GNOME-p3}.}, and can be used to conduct novelty-driven experiments. In the DARPA SAIL-ON program, we do so by first plugging in an externally developed agent (trained using, for example, reinforcement learning techniques) that has been designed by the external team to detect and react to novelty. The other three agents, at present, are combinations of the heuristic agents that were used in the demonstration, and are also available on GitHub. These agents are non-adaptive, but their heuristics cover a fairly comprehensive and established set of strategies (for playing Monopoly `well') and their actions are interpretable. 

Before describing the experimental protocol briefly, we note that two engineering challenges that we had to address during development of the simulator was the \emph{avoidance of race conditions} (since, in the real game, players could simultaneously be making decisions, which may conflict) and \emph{sufficient modularity} so that novelty could be injected while minimizing the chances of an agent `crashing' due to \emph{syntactic} causes. Software engineering details behind the simulator, and its core design principles, are described in an article currently under review\footnote{For the interested reviewer, we provide a preliminary draft (not to be circulated) here: \url{https://drive.google.com/file/d/1gGy9lmiF91H735PcODetPH7_M9HAAhRV/view?usp=sharing}}. 

To ensure the experiments test the agent's reaction to novelty that is truly unanticipated (and hence, would be the closest proxy for expected agent reaction to novelty in the open world), the novelty generator used during evaluations is not revealed publicly. However, an example novelty generator is available in the GitHub repository for teams to self-evaluate on. In the demonstration, we used a subset of novelties from this generator. Many other novelties are supported by the publicly available generator, in addition to the ones used in the demonstration.  

For each novelty, during test-time, the agent has to play, not a single game of Monopoly, but a sequence of games (called a \emph{tournament}). For a preset, tunable parameter $k$ (that is unknown to the agent developing team and that can vary in a narrow range in each tournament), the first $k$ games of the tournament constitute the \emph{pre-novelty phase} and are used to record the \emph{default} performance of the agent. In current experiments, we use a performance metric called the Win Ratio, which simply records the fraction of games won by the agent. Starting from $k$, the novelty is injected at the beginning of every game, till the final game of the tournament. 

For example, if the length of the tournament was 40 games, and $k$ was 10, the first 9 games would be run without novelty, and games 10-40 would have novelty injected into them. The notion is that the world `changes' at game 10 (and the \emph{post-novelty phase} begins), and the agent must not only detect the novelty (which is sometimes obvious, as in the novelties used in the demonstration, but can also be statistically subtle, such as if one of the die became heavily biased, rather than obey a uniform distribution) but also react to it. Since it may take a few games for the agent to either detect or react to the novelty, we record not just the mean Win Ratio over the entire post-novelty phase but also the Win Ratio toward the end of the tournament. The idea is that such an `asymptotic' Win Ratio would capture the agent's maximal reaction capacity to that novelty.

In conducting the experiments above, the standard experimental protocols are followed. Due to stochasticity, multiple tournaments are always run, even for a single novelty. Statistical significance and standard errors are always reported in evaluation brief-outs. We use a library of approximately 45-50 novelties for our evaluation, spread across class, attribute and representation novelty categories, and a range of expected degrees of difficulty. In addition to recording and reporting the magnitude of the percentage change in Win Ratio between the pre-novelty and post-novelty phases as the novelty reaction performance, the SAIL-ON program also requires agents to emit a binary signal when they are confident that they have detected the novelty, which is used to evaluate whether (and how quickly) an agent was able to detect that novelty. 

To encourage robust novelty adaptation, the novelty injected at the beginning of a game may not be static but could be from a distribution. For example, if the novelty we were evaluating on was the dice novelty described earlier for the demonstration, we may uniformly sample a number between 3-5 and inject the sampled number of dice at the beginning of the game. This number could obviously change from game to game, but the distribution stays fixed during the post-novelty phase of that tournament.     

Perhaps most importantly, we note that an element of good faith is always necessary in such evaluations. Agents developed by performers in the SAIL-ON program are expected to not deliberately try to use syntactic or other cues to either detect or `play' around the novelty. Instead, they may only rely on their prior model of the game, as well as publicly observable information during gameplay, including actions that other agents may be taking. Agents are also expected to be reasonably robust. To minimize syntactic issues, we typically have a bake-in period before the formal evaluation, where we attempt to run the agent model and verify that it does not crash. Periodically, we also release small sets of `unanticipated' novelties used in evaluations to help the teams gain insights and improve novelty reaction and detection performance in the next evaluation on the remainder of the (unrevealed) novelties.

\section{Future Work}

There are many areas of future work worth exploring. The most interesting of these is into the nature of novelty itself, including the theoretical strengths and limits of the operational definition stated in Section \ref{novelty}, and also how the theory complies (or doesn't comply) with \emph{human} intuitions of novelty. We are continuing to implement more advanced types of novelty in the novelty generator in GNOME; particularly, \emph{interaction} novelties, wherein agents are capable of a broader set of interactions, including collusion, which would allow the implementation of the game to  more accurately mirror some of the more interesting and `open-world' properties of Monopoly when human beings play it for recreation. We are also investigating \emph{environment} novelties e.g., what if an AI agent trained on a gameboard designed for the US market now has to play the version designed for the UK market? 

Another interesting category of novelties are at the level of \emph{game rules}. These novelties are interesting because, at the extreme, they may lead to a situation where the game is very different from the default game, and the injection of the novelty has (in essence) led to a change in \emph{domain}. We hope that investigation and evaluation of these novelties will lead to insights into the distinctions between the \emph{fundamental} and \emph{malleable} aspects of a chosen domain. The hypothesis is that the novelties concerned with the former aspects will somehow be easier to adapt to than the novelties concerned with the latter. More generally, we expect that these more advanced novelties, and evaluation of AI agents in real-time when these novelties are injected, will lead to both theoretical and empirical insights into the nature of novelty and its intersection with an agent's learning capabilities. 

Finally, a longer-term goal is to implement other multi-agent games, such as Poker, with novelty generators, to support a general experimental framework that treats novelty as a first-class citizen.

\section{Conclusion} 

In this article, we briefly described a simulator called GNOME (Generating Novelty in Open-World Multi-agent Environments) that has been developed to support the development, training and evaluation of agents that can detect and react to \emph{novelty}, defined operationally in the SAIL-ON program as the  `states or situations that violate (implicit or explicit) assumptions about agents, the environment, and agent-agent and agent-environment interactions.' GNOME is meant to support this goal in a domain-specific manner by permitting controlled injection of novelty in the game of Monopoly. We have designed the the simulator to be extensible and modular, with simple decision interfaces that allow agents to be plugged into the simulator, a novelty generator that is delineated from the overall simulator architecture and logic, and that can therefore stay unrevealed to the agent developer, allowing development of agent algorithms and models that could react to \emph{unanticipated} novelty. GNOME was recently demonstrated through a web interface at the NeurIPS 2020 conference, where participants could inject novelties from a wide range of options, select the AI players, and observe how the game-state evolved in the face of those novelties. 
% \newpage
% Acknowledgements should go at the end, before appendices and references

\acks{This work was funded by the Defense Advanced Research Projects Agency (DARPA) with award W911NF2020003 under the SAIL-ON program. }


% \newpage

% \appendix
% \section*{Appendix A.}
% \label{app:theorem}

% % Note: in this sample, the section number is hard-coded in. Following
% % proper LaTeX conventions, it should properly be coded as a reference:

% %In this appendix we prove the following theorem from
% %Section~\ref{sec:textree-generalization}:

% In this appendix we prove the following theorem from
% Section~6.2:

% \noindent
% {\bf Theorem} {\it Let $u,v,w$ be discrete variables such that $v, w$ do
% not co-occur with $u$ (i.e., $u\neq0\;\Rightarrow \;v=w=0$ in a given
% dataset $\dataset$). Let $N_{v0},N_{w0}$ be the number of data points for
% which $v=0, w=0$ respectively, and let $I_{uv},I_{uw}$ be the
% respective empirical mutual information values based on the sample
% $\dataset$. Then
% \[
% 	N_{v0} \;>\; N_{w0}\;\;\Rightarrow\;\;I_{uv} \;\leq\;I_{uw}
% \]
% with equality only if $u$ is identically 0.} \hfill\BlackBox

% \noindent
% {\bf Proof}. We use the notation:
% \[
% P_v(i) \;=\;\frac{N_v^i}{N},\;\;\;i \neq 0;\;\;\;
% P_{v0}\;\equiv\;P_v(0)\; = \;1 - \sum_{i\neq 0}P_v(i).
% \]
% These values represent the (empirical) probabilities of $v$
% taking value $i\neq 0$ and 0 respectively.  Entropies will be denoted
% by $H$. We aim to show that $\fracpartial{I_{uv}}{P_{v0}} < 0$....\\

% {\noindent \em Remainder omitted in this sample. See http://www.jmlr.org/papers/ for full paper.}

\vskip 0.2in


\begin{thebibliography}{25}
\providecommand{\natexlab}[1]{#1}
\providecommand{\url}[1]{\texttt{#1}}
\expandafter\ifx\csname urlstyle\endcsname\relax
  \providecommand{\doi}[1]{doi: #1}\else
  \providecommand{\doi}{doi: \begingroup \urlstyle{rm}\Url}\fi

\bibitem[Abu-Mostafa(1989)]{ML5}
Yaser~S Abu-Mostafa.
\newblock The vapnik-chervonenkis dimension: Information versus complexity in
  learning.
\newblock \emph{Neural Computation}, 1\penalty0 (3):\penalty0 312--317, 1989.

\bibitem[Arulkumaran et~al.(2017)Arulkumaran, Deisenroth, Brundage, and
  Bharath]{arulkumaran2017deep}
Kai Arulkumaran, Marc~Peter Deisenroth, Miles Brundage, and Anil~Anthony
  Bharath.
\newblock Deep reinforcement learning: A brief survey.
\newblock \emph{IEEE Signal Processing Magazine}, 34\penalty0 (6):\penalty0
  26--38, 2017.

\bibitem[Bhaskar(2002)]{phil1}
Roy Bhaskar.
\newblock The philosophy of meta-reality: Part ii: Agency, perfectibility,
  novelty.
\newblock \emph{Journal of critical realism}, 1\penalty0 (1):\penalty0 67--93,
  2002.

\bibitem[Brockman et~al.(2016)Brockman, Cheung, Pettersson, Schneider,
  Schulman, Tang, and Zaremba]{openaigym}
Greg Brockman, Vicki Cheung, Ludwig Pettersson, Jonas Schneider, John Schulman,
  Jie Tang, and Wojciech Zaremba.
\newblock Openai gym.
\newblock \emph{arXiv preprint arXiv:1606.01540}, 2016.

\bibitem[Brown and Sandholm(2019)]{poker1}
Noam Brown and Tuomas Sandholm.
\newblock Superhuman ai for multiplayer poker.
\newblock \emph{Science}, 365\penalty0 (6456):\penalty0 885--890, 2019.

\bibitem[Cesa-Bianchi et~al.(2004)Cesa-Bianchi, Conconi, and Gentile]{ML4}
Nicolo Cesa-Bianchi, Alex Conconi, and Claudio Gentile.
\newblock On the generalization ability of on-line learning algorithms.
\newblock \emph{IEEE Transactions on Information Theory}, 50\penalty0
  (9):\penalty0 2050--2057, 2004.

\bibitem[Chandola et~al.(2009)Chandola, Banerjee, and Kumar]{anomalydetection}
Varun Chandola, Arindam Banerjee, and Vipin Kumar.
\newblock Anomaly detection: A survey.
\newblock \emph{ACM computing surveys (CSUR)}, 41\penalty0 (3):\penalty0 1--58,
  2009.

\bibitem[Chu and Macgregor(2011)]{chu2011}
Yun Chu and James Macgregor.
\newblock Human performance on insight problem solving: A review.
\newblock \emph{The Journal of Problem Solving}, 3\penalty0 (2):\penalty0
  119--150, 2011.

\bibitem[Cruz(2019)]{diplomacy}
Diogo Henrique~Marques Cruz.
\newblock Deep reinforcement learning in strategic multi-agent games: the case
  of no-press diplomacy.
\newblock 2019.

\bibitem[Gama et~al.(2014)Gama, {\v{Z}}liobait{\.e}, Bifet, Pechenizkiy, and
  Bouchachia]{conceptdrift}
Jo{\~a}o Gama, Indr{\.e} {\v{Z}}liobait{\.e}, Albert Bifet, Mykola Pechenizkiy,
  and Abdelhamid Bouchachia.
\newblock A survey on concept drift adaptation.
\newblock \emph{ACM computing surveys (CSUR)}, 46\penalty0 (4):\penalty0 1--37,
  2014.

\bibitem[Ho and Ermon(2016)]{ho2016generative}
Jonathan Ho and Stefano Ermon.
\newblock Generative adversarial imitation learning.
\newblock In \emph{Advances in neural information processing systems}, pages
  4565--4573, 2016.

\bibitem[Laptev et~al.(2015)Laptev, Amizadeh, and Flint]{TS}
Nikolay Laptev, Saeed Amizadeh, and Ian Flint.
\newblock Generic and scalable framework for automated time-series anomaly
  detection.
\newblock In \emph{Proceedings of the 21th ACM SIGKDD international conference
  on knowledge discovery and data mining}, pages 1939--1947, 2015.

\bibitem[Michie et~al.(1994)Michie, Spiegelhalter, Taylor, et~al.]{ML2}
Donald Michie, David~J Spiegelhalter, CC~Taylor, et~al.
\newblock Machine learning.
\newblock \emph{Neural and Statistical Classification}, 13\penalty0
  (1994):\penalty0 1--298, 1994.

\bibitem[Mnih et~al.(2016)Mnih, Badia, Mirza, Graves, Lillicrap, Harley,
  Silver, and Kavukcuoglu]{atari}
Volodymyr Mnih, Adria~Puigdomenech Badia, Mehdi Mirza, Alex Graves, Timothy
  Lillicrap, Tim Harley, David Silver, and Koray Kavukcuoglu.
\newblock Asynchronous methods for deep reinforcement learning.
\newblock In \emph{International conference on machine learning}, pages
  1928--1937, 2016.

\bibitem[Nadeau and Bengio(2003)]{ML1}
Claude Nadeau and Yoshua Bengio.
\newblock Inference for the generalization error.
\newblock \emph{Machine learning}, 52\penalty0 (3):\penalty0 239--281, 2003.

\bibitem[Pathak et~al.(2017)Pathak, Agrawal, Efros, and
  Darrell]{pathak2017curiosity}
Deepak Pathak, Pulkit Agrawal, Alexei~A Efros, and Trevor Darrell.
\newblock Curiosity-driven exploration by self-supervised prediction.
\newblock In \emph{Proceedings of the IEEE Conference on Computer Vision and
  Pattern Recognition Workshops}, pages 16--17, 2017.

\bibitem[Pimentel et~al.(2014)Pimentel, Clifton, Clifton, and
  Tarassenko]{pimentel2014}
Marco~AF Pimentel, David~A Clifton, Lei Clifton, and Lionel Tarassenko.
\newblock A review of novelty detection.
\newblock \emph{Signal Processing}, 99:\penalty0 215--249, 2014.

\bibitem[Shao et~al.(2009)Shao, Xie, Hong, and Jost]{fraud}
Xuhui Shao, Jianjun Xie, Tao Hong, and Allen Jost.
\newblock System and method for identity-based fraud detection through graph
  anomaly detection, July~21 2009.
\newblock US Patent 7,562,814.

\bibitem[Silver et~al.(2017)Silver, Schrittwieser, Simonyan, Antonoglou, Huang,
  Guez, Hubert, Baker, Lai, Bolton, et~al.]{alphago}
David Silver, Julian Schrittwieser, Karen Simonyan, Ioannis Antonoglou, Aja
  Huang, Arthur Guez, Thomas Hubert, Lucas Baker, Matthew Lai, Adrian Bolton,
  et~al.
\newblock Mastering the game of go without human knowledge.
\newblock \emph{nature}, 550\penalty0 (7676):\penalty0 354--359, 2017.

\bibitem[Silver et~al.(2018)Silver, Hubert, Schrittwieser, Antonoglou, Lai,
  Guez, Lanctot, Sifre, Kumaran, Graepel, et~al.]{alphazero}
David Silver, Thomas Hubert, Julian Schrittwieser, Ioannis Antonoglou, Matthew
  Lai, Arthur Guez, Marc Lanctot, Laurent Sifre, Dharshan Kumaran, Thore
  Graepel, et~al.
\newblock A general reinforcement learning algorithm that masters chess, shogi,
  and go through self-play.
\newblock \emph{Science}, 362\penalty0 (6419):\penalty0 1140--1144, 2018.

\bibitem[Simon and Newell(1971)]{simon1971}
Herbert~A Simon and Allen Newell.
\newblock Human problem solving: The state of the theory in 1970.
\newblock \emph{American Psychologist}, 26\penalty0 (2):\penalty0 145--159,
  1971.

\bibitem[Tang et~al.(2018)Tang, Shao, Zhu, Li, Zhao, and Huang]{agent2}
Zhentao Tang, Kun Shao, Yuanheng Zhu, Dong Li, Dongbin Zhao, and Tingwen Huang.
\newblock A review of computational intelligence for starcraft ai.
\newblock In \emph{2018 IEEE Symposium Series on Computational Intelligence
  (SSCI)}, pages 1167--1173. IEEE, 2018.

\bibitem[Vinyals et~al.(2019)Vinyals, Babuschkin, Czarnecki, Mathieu, Dudzik,
  Chung, Choi, Powell, Ewalds, Georgiev, et~al.]{starcraft1}
Oriol Vinyals, Igor Babuschkin, Wojciech~M Czarnecki, Micha{\"e}l Mathieu,
  Andrew Dudzik, Junyoung Chung, David~H Choi, Richard Powell, Timo Ewalds,
  Petko Georgiev, et~al.
\newblock Grandmaster level in starcraft ii using multi-agent reinforcement
  learning.
\newblock \emph{Nature}, 575\penalty0 (7782):\penalty0 350--354, 2019.

\bibitem[Wisdom(1944)]{phil2}
John Wisdom.
\newblock Philosophy, anxiety and novelty.
\newblock \emph{Mind}, 53\penalty0 (210):\penalty0 170--176, 1944.

\bibitem[Zhang et~al.(2016)Zhang, Bengio, Hardt, Recht, and Vinyals]{ML3}
Chiyuan Zhang, Samy Bengio, Moritz Hardt, Benjamin Recht, and Oriol Vinyals.
\newblock Understanding deep learning requires rethinking generalization.
\newblock \emph{arXiv preprint arXiv:1611.03530}, 2016.

\end{thebibliography}
\end{document}